\documentclass[runningheads]{llncs}

\usepackage{eccv}

\usepackage{eccvabbrv}

\usepackage{graphicx}
\usepackage{booktabs}

\usepackage[accsupp]{axessibility}  

\usepackage[pagebackref,breaklinks,colorlinks]{hyperref}

\usepackage{orcidlink}

\usepackage[misc,geometry]{ifsym}

\usepackage[dvipsnames]{xcolor}
\usepackage{colortbl}
\usepackage{multirow}
\usepackage{amssymb}
\usepackage{amsmath}
\usepackage{wrapfig}
\usepackage{float}

\usepackage{amsmath}
\usepackage{amsfonts}
\usepackage{amssymb}
\usepackage{wrapfig}
\usepackage{subcaption}
\usepackage{multirow}
 \usepackage{mathtools} 

\usepackage{verbatim}

\usepackage{anyfontsize}

\usepackage{microtype}
\usepackage{graphicx}
\usepackage{booktabs} 
\usepackage{multirow}
\usepackage{amsmath,amssymb}
\usepackage{booktabs}
\usepackage{caption,subcaption}

\usepackage{xcolor}
\definecolor{mygreen}{HTML}{3cb44b}
\definecolor{skyblue}{HTML}{beffff}
\definecolor{lightgreen}{HTML}{90ee90}

\usepackage{color, colortbl}

\definecolor{emerald}{rgb}{0.31, 0.78, 0.37}

\usepackage{tcolorbox}
\usepackage{enumitem}
\setitemize{itemsep=10pt,topsep=0pt,parsep=0pt,partopsep=0pt}
\usepackage{colortbl}

\usepackage{xcolor}
\definecolor{mygreen}{HTML}{3cb44b}
\colorlet{myyellow}{green!10!orange!90!}
\definecolor{myblue}{rgb}{0.36, 0.61, 0.84}
\definecolor{myorange}{rgb}{0.93, 0.49, 0.19}
\makeatletter

\usepackage{tikz}
\usetikzlibrary{arrows,shapes,snakes,automata,backgrounds,fit,petri}
\usepackage{adjustbox}

\newcommand{\RN}[1]{%
	\textup{\lowercase\expandafter{\it \romannumeral#1}}%
}
\usepackage{tabu}

\newcommand{\beq}{\vspace{0mm}\begin{equation}}
\newcommand{\eeq}{\vspace{0mm}\end{equation}}
\newcommand{\beqs}{\vspace{0mm}\begin{eqnarray}}
\newcommand{\eeqs}{\vspace{0mm}\end{eqnarray}}
\newcommand{\barr}{\begin{array}}
\newcommand{\earr}{\end{array}}

\usepackage{color, colortbl}
\definecolor{Gray}{gray}{0.93}

\usepackage{lipsum}

\usepackage{pifont}
\newcommand{\cmark}{\ding{51}}%
\newcommand{\xmark}{\ding{55}}%

\usepackage{makecell}

\usepackage{xcolor,amsmath}
\usepackage[linesnumbered,ruled,vlined]{algorithm2e}
\DontPrintSemicolon

\usepackage{xcolor}
\definecolor{mygreen}{HTML}{3cb44b}

\SetKwComment{Comment}{\color{green!50!black}\# }{}

\SetKwProg{Function}{def}{:}{}

\SetKwProg{For}{for}{:}{}
\SetKwProg{If}{if}{:}{}

\newcommand{\PredSty}[1]{\textnormal{\ttfamily\color{mygreen!90!black}#1}\unskip}

\def\CircleArrowright{\ensuremath{%
  \rotatebox[origin=c]{310}{$\circlearrowright$}}}
\DeclareRobustCommand{\star}{%
  \begingroup\normalfont
  \includegraphics[height=\fontcharht\font`\B]{figures/star.pdf}%
  \endgroup
}
\newcommand{\vlnbert}{VLN$\protect\CircleArrowright$BERT}

\newcommand{\ours}{NavGPT-2}

\begin{document}

\title{\ours: Unleashing Navigational Reasoning Capability for Large Vision-Language Models}

\titlerunning{\ours}

\author{Gengze Zhou\inst{1}\orcidlink{0000-0003-0279-9277} \and
Yicong Hong\inst{2}\orcidlink{0000-0002-5068-1508} \and
Zun Wang\inst{3}\orcidlink{0009-0005-9502-050X} \and
\\ Xin Eric Wang\inst{4}\orcidlink{0000-0003-2605-5504} \and
Qi Wu\inst{1}$^{\left(\text{\Letter}\right)}$\orcidlink{0000-0003-3631-256X}}

\authorrunning{G.~Zhou et al.}

\institute{AIML, University of Adelaide, Adelaide, Australia\\
\email{\{gengze.zhou, qi.wu01\}@adelaide.edu.au} \and
Adobe Research, San Jose, USA \and
University of North Carolina, Chapel Hill, USA \and
University of California, Santa Cruz, USA }

\maketitle

\begin{abstract}
  Capitalizing on the remarkable advancements in Large Language Models (LLMs), there is a burgeoning initiative to harness LLMs for instruction following robotic navigation. Such a trend underscores the potential of LLMs to generalize navigational reasoning and diverse language understanding. However, a significant discrepancy in agent performance is observed when integrating LLMs in the Vision-and-Language navigation (VLN) tasks compared to previous downstream specialist models. Furthermore, the inherent capacity of language to interpret and facilitate communication in agent interactions is often underutilized in these integrations. In this work, we strive to bridge the divide between VLN-specialized models and LLM-based navigation paradigms, while maintaining the interpretative prowess of LLMs in generating linguistic navigational reasoning. By aligning visual content in a frozen LLM, we encompass visual observation comprehension for LLMs and exploit a way to incorporate LLMs and navigation policy networks for effective action predictions and navigational reasoning. We demonstrate the data efficiency of the proposed methods and eliminate the gap between LM-based agents and state-of-the-art VLN specialists. The source code is available at \url{https://github.com/GengzeZhou/NavGPT-2}.
  \keywords{Vision-and-Language Navigation \and Large Language Models \and Vision-Language Models}
\end{abstract}

\section{Introduction}

Motivating by the considerable advances in Large Language Models (LLMs), there is an emerging effort to utilize these models for instructional tasks within robotic navigation~\cite{zhou2024navgpt, long2023discuss, pan2023langnav, chen2024mapgpt, zheng2023towards}. This development highlights two core capacities of LLMs: Firstly, the ability to generalize commonsense knowledge reasoning and efficiently process free-form linguistic inputs, thanks to learning enormous amounts of textual data from the web. Secondly, the interpretative of LLMs to provide navigational reasoning explicitly in a human interpretable way and the associated communicative potential during interaction with humans. 
Several studies have been initiated to integrate LLMs in the context of Vision-and-Language Navigation (VLN)~\cite{anderson2018r2r}. Specifically, we recognize two major lines of research, including \textit{zero-shot VLN with LLMs} and \textit{finetune LLMs for VLN}. However, these approaches reveal a notable performance gap towards agents designed and trained tailored for solving VLN~\cite{zhou2024navgpt, long2023discuss, chen2024mapgpt}, usually lie at two extremes that carry significant limitations: 

\begin{figure}[t]
    \centering
    \includegraphics[width=0.98\linewidth]{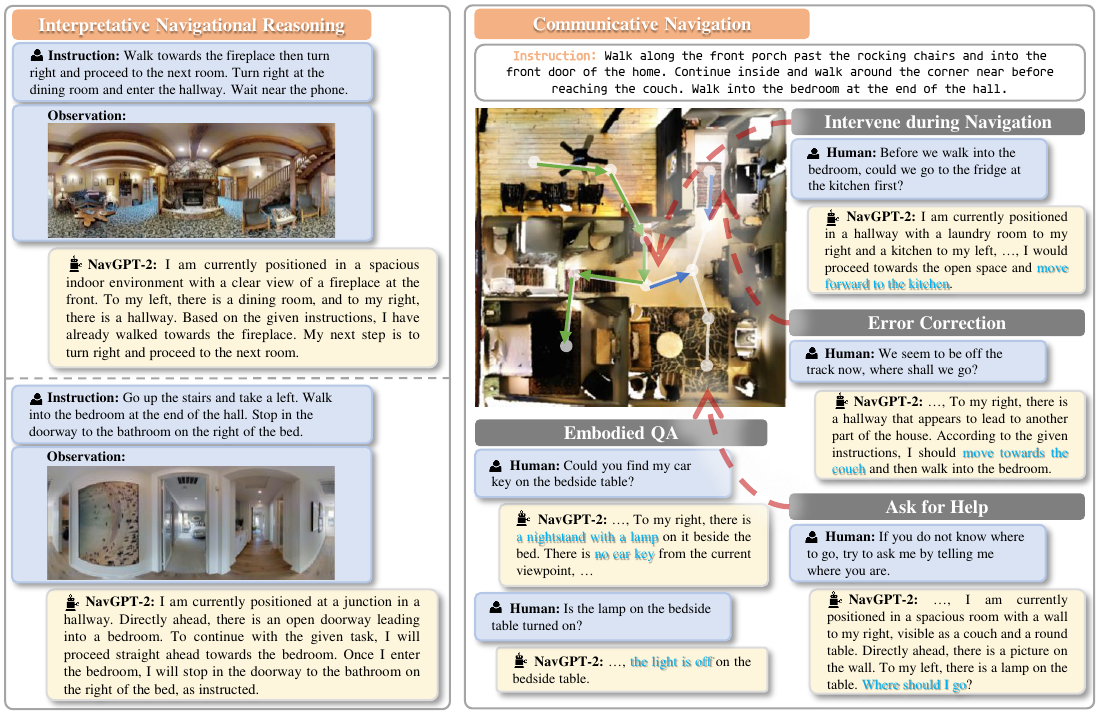}
    \caption{Left: Besides performing effective navigation planning, \ours\ is capable of generating navigational reasoning in a human-interpretable way. Right: \ours\ can support multi-round interaction with the user and plan according to the user’s intervention in the navigation process, actively ask for help, and answer visual questions.
    }
    \label{fig:reasoning}
\end{figure}

-- For the zero-shot approach, previous studies prompt the LLMs with comprehensive descriptions of the navigation task and progress on the fly as the agent moves in an environment~\cite{zhou2024navgpt,chen2024mapgpt,long2023discuss, zhan2024mc, long2024instructnav}. To enable the language models to ``see'', the agent's stepwise observations are often translated to textual descriptions generated by image captioning models such as BLIP-2~\cite{li2023blip2}. Meanwhile, the agent's past observations and decisions are summarized in pure language because the partially observable VLN process demands the agent to keep track of their experience for future planning. Eventually, the LLMs will read all the above information and infer an action that takes the agent towards the target. Despite the fact that such formulation largely exploits the generalizable knowledge of LLMs for navigation and eliminates the need to train a specialized agent on scarce embodied data, it involves complicated and fragile prompt engineering, increasingly expensive step-wise prompting, as well as noisy captioning and summarization which inevitably causes loss of information. More crucially, it still remains an open question whether LLMs can correctly understand spatial structures and the consequence of physical motions, as evident by the huge performance gap between the LLM zero-shot approach and VLN-specialized models ($\sim$40\% success rate gap on R2R~\cite{anderson2018r2r}). As a result, we suggest that the zero-shot approach based on existing LLMs is likely infeasible to solve VLN.

\setcounter{footnote}{0}
-- For the fine-tuning approach, a range of research attempts to tune LLMs on instruction-trajectory pairs from VLN datasets~\cite{pan2023langnav, zheng2023towards, zhang2024navid, lin2024navcot}. 
Specifically, the visual observations are fed to the language model either as encoded representations or as textual descriptions, and the actions are generally converted to structured textual format to adapt the auto-regressive prediction training of language models~\cite{pan2023langnav, zhang2024navid}. 
Although these approaches~\cite{pan2023langnav, lin2024navcot, zheng2023towards} utilize the generalizable pretrained weights of LLMs and are much larger in scale\footnote{LLaMA2-7B~\cite{touvron2023llama2} have 6.74 billion parameters while DUET~\cite{chen2022duet} has only 0.18 billion.}, their performance still falls far behind the VLN-specialized models, likely due to the insufficient amount of training data and the discrepancy between LLMs' pretraining objectives and VLN's training target of aligning multi-view images and partial instruction.
More crucially, directly tuning LLMs for VLN discards their general language capabilities, for instance, the potential to describe and explain the navigation process and communicate with humans for interactive tasks. Losing these abilities in fact against one of the most important motivations of introducing LLMs to embodied AI, yielding ``black-box'' uncontrollable agents.  

In light of this, we propose \ours{}, a system that finds a balance between the two aforementioned extremes, incorporating effective navigational modules to facilitate navigational capabilities for VLM. Specifically, we built upon the VLM architecture of InstructBLIP~\cite{dai2023instructblip}, facilitating it with multi-image perception to adapt to VLN tasks. We enable VLM with navigational reasoning ability by constructing step-wise navigational reasoning data with GPT-4V and performing visual instruction tuning. We employ VLM latent as the composite visual-linguistic representation for both language decoding and action decoding with topological graphs, to enable agents to trace the long-term navigation history and effective backtracking while preserving the general language abilities of LLMs. As shown in Figure \ref{fig:reasoning}, \ours\ could generate interpretative actions with language, and demonstrate the significant potential of building a communicative VLN agent that allows users to receive feedback and foster a connection with robots.

Our contributions are as follows: (1) We propose a pipeline to incorporate VLN specialists with VLMs free from LLM training. (2) Leveraging the robust feature enhancement afforded by pretrained VLMs, \ours\ eliminates the gap between LM-based agents and SOTA VLN specialists. (3) We reserve the communicative instinct of LMs, enabling the models to explicitly explain the reason behind each navigation decision. These abilities are essential for building a practical and interactive VLN agent.

\section{Related Works}
\noindent\textbf{\emph{Vision-and-Language Navigation (VLN)}}
The pursuit of developing a universal navigation agent capable of following free-form linguistic directives to navigate within an unfamiliar photorealistic environment has been a longstanding objective in the field of Vision-and-Language Navigation~\cite{anderson2018vision, anderson2020rxr, qi2020reverie, thomason2020cvdn}.
These methods solve the task from two main aspects.
(1) Vision-Language alignment. Some work~\cite{hao2020prevalent, hong2020recurrent, guhur2021airbert, majumdar2020improving, chen2021hamt, chen2022duet, qiao2023hop+, li2019robust, zhao2023mind} benefits from generic visual-linguistic representations~\cite{chen2020uniter, li2020oscar, su2019vl, li2019visualbert, tan2019lxmert}, 
some exploit additional supervision from data augmentation~\cite{anderson2018spl, tan2019envdrop, li2022envedit, parvaneh2020counterfactual, wang2023scaling, guhur2021airbert, dou2022foam, wang2022counterfactual, li2023panogen}, 
training stragies~\cite{wang2018look, wang2019reinforced, ma2019self, zhu2020vision, huang2019transferable} to learn such cross-modality alignment.
(2) Efficient action planning mechanism with historical state memorization~\cite{hong2020recurrent, chen2021history}, 
self-correction~\cite{ke2019tactical, ma2019regretful, zhao2023mind}, 
navigation map construction~\cite{an2021neighbor, chen2021topological, chen2022think, wang2020active, zhao2022target, liu2023bird, wang2023gridmm}, 
and external knowledge prompts~\cite{lin2022adapt, li2023kerm}.
In this work, we investigate the method of boosting a simplified prevalent VLN policy with visual-linguistic representations in LLM's latent space.

\noindent\textbf{\emph{Large Language Models in VLN}}
Introducing Large Language Models (LLMs) in robotics navigation is for its superior language understanding, communicative intrinsic, and commonsense reasoning ability. Previous work in VLN mainly exploits three strategies to combine LLM in solving VLN: (1) Model ensemble. Mic~\cite{qiao2023march} utilizes GPT-2~\cite{radford2019language} to provide extra guidance for downstream VLN specialists. (2) LLM as zero-shot VLN agents. NavGPT~\cite{zhou2024navgpt} reveals the potential navigational reasoning capacity of off-the-shell LLMs with a complex prompting system. DiscussNav~\cite{long2023discuss} propose a multi-agent system while MapGPT~\cite{chen2024mapgpt} introduce topological mapping for zero-shot VLN agents to further improve their performance. However, a large performance gap is observed compared to supervised methods, even if the most powerful GPT-4~\cite{openai2023gpt4} models are used. (3) Finetune language models as VLN agents. LangNav~\cite{pan2023langnav} and NavCoT~\cite{lin2024navcot} finetune a LlaMA-7B~\cite{touvron2023llama} on VLN data to investigate the effectiveness of language as perceptual representation to perform navigation. NaviLLM~\cite{zheng2023towards} perform multi-task learning to finetune a LlaMA-7B~\cite{touvron2023llama} into VLN generalist.

\noindent\textbf{\emph{Modality Alignment in Large Vision-Language Models (VLM)}} 
Leveraging the recent progress in Large Language Models (LLMs), there is a growing endeavor to repurpose pretrained Large Language Models for multimodal tasks, encompassing the comprehension and interpretation of visual information~\cite{alayrac2022flamingo, liu2024visual, li2023blip2, dai2023instructblip, driess2023palm, zhu2023minigpt, chen2023minigpt, peng2023kosmos, wang2024visionllm, bai2023qwen}. A feasible way to connect different modalities for the pretrained LLMs is by training query-based visual resamplers, which is initially introduced by Flamingo~\cite{alayrac2022flamingo} and BLIP-2~\cite{li2023blip2}, and followed by subsequent implementations in MiniGPT-4~\cite{zhu2023minigpt}, InstructBLIP~\cite{dai2023instructblip} and Qwen-VL~\cite{bai2023qwen}. Concurrently, another strand of research has focused on employing fully connected projection layers to directly map the output from vision encoders to the input of LLMs. This method is exemplified by LLaVA~\cite{liu2024visual} and MiniGPT-v2~\cite{chen2023minigpt}. In this work, we adopt the Q-former designed in InsturctBLIP~\cite{dai2023instructblip} to effectively control the length of visual content for multiple view images input at each viewpoint.

\section{Method}

\begin{figure}[t]
    \centering
    \includegraphics[width=.99\linewidth]{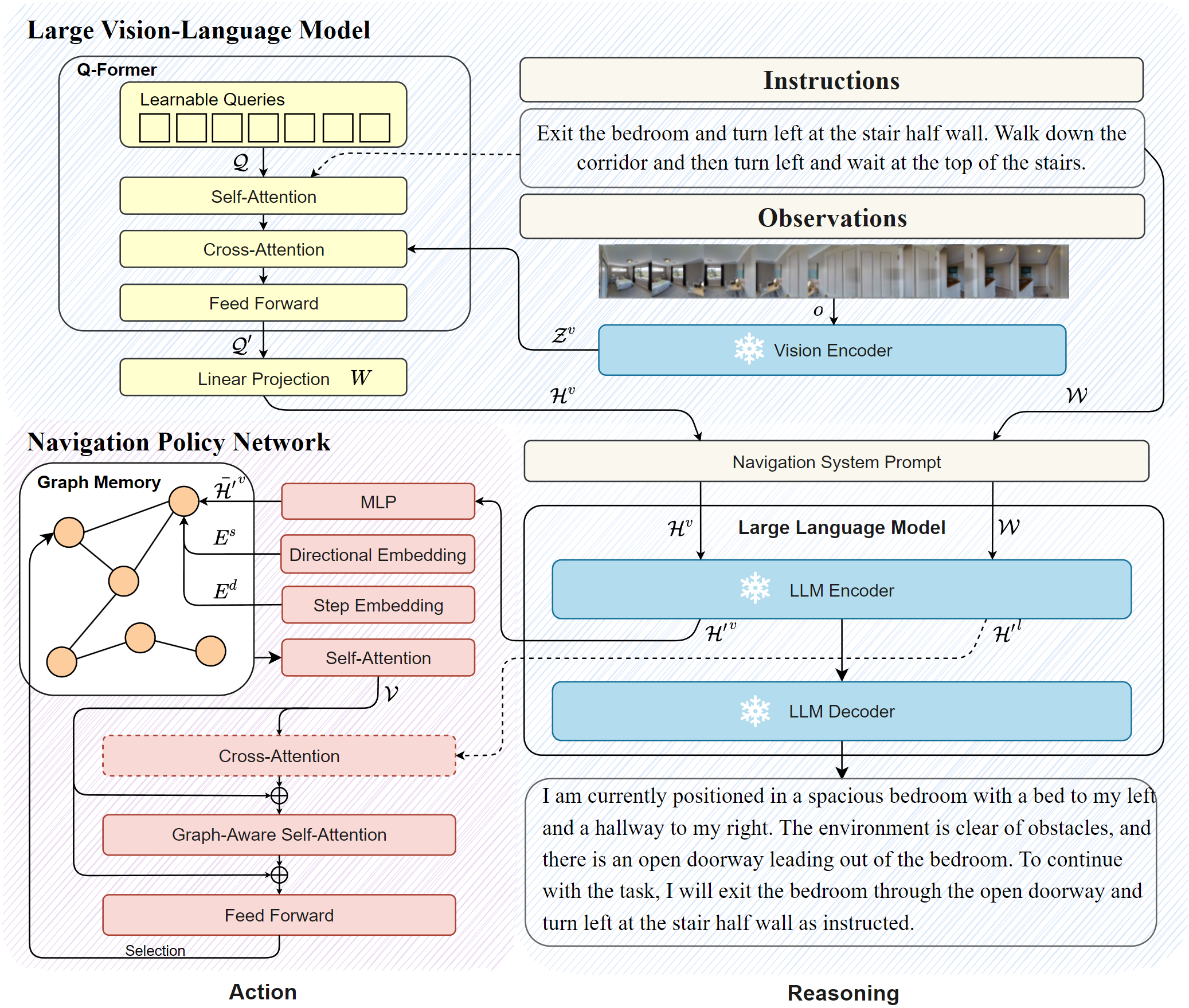}
    \caption{Model architecture of \ours, it consists of a multimodality Large Language Model and a topological graph-based navigation policy network. The yellow blocks indicate the trainable module at stage one, the red blocks indicate the trainable module at stage two, and the blue blocks are frozen.}
    \label{fig:architecture}
\end{figure}

The architecture of \ours, as depicted in Figure~\ref{fig:architecture}, comprises two primary components: a Large Vision-Language Model (VLM) and a navigation policy network. Within the VLM, visual observations and instructions are processed by a component referred to as the Q-former to extract image tokens. These tokens serve as the input visual content for the LLM, enabling it to generate navigational reasoning. For action prediction, the model employs both hidden representations of image tokens and instruction text tokens that have been processed by the LLM encoder as the input features.

\subsubsection{\emph{Problem Formulation.}} Given an instruction composed of $L$ word embeddings $\mathcal{W} = \{w_i\}^L_{i=1}$, an agent is required to follow the instruction to navigate in a pre-defined undirected graph $\mathcal{G} = \{\mathcal{V}, \mathcal{E}\}$, where $\mathcal{V}$ denotes the navigable nodes, $\mathcal{E}$ denotes the connectivity edges. At step $t$, the agent perceives the surrounding environment through the observation of a set of RBG views for each connected navigable node candidate $\mathcal{O}_t \triangleq \{\langle o_i, a_i \rangle\}^N_{i=1}$, where $N$ denotes the number of candidate nodes, each unique view is denoted as $o_i (i\leq N)$, with its associated angle direction with respect to the agent's heading represented as $a_i (i \leq N)$. The agent predicts the subsequent action by selecting the relative angle $a_t$ from $\mathcal{O}_t$, the policy $\pi$ parametrized by ${\Theta}$ that the agent is required to learn is $\pi(a_t|\mathcal{W}, \mathcal{O}_t; {\Theta})$.

\subsection{VLMs Latent as Visual-Linguistic Representation}
In this section, we discuss the model design within the Large Vision-Language Model, how to enable frozen LLMs to generate navigational reasoning, and how to utilize VLM features in the navigation policy network.
\subsubsection{\emph{Visual Aligning with LLMs.}}
To effectively encode multiple-view images in the environment and construct spatial perception for navigation reasoning in a frozen LLM, we adopt the Q-former~\cite{li2023blip2} design and encode each view into fixed-length visual tokens, shown in Figure~\ref{fig:architecture}. Specifically, for a candidate view image $o_i$, we incorporate a frozen ViT-g/14 from EVA-CLIP~\cite{fang2023eva} as the vision encoder to extract visual feature $\mathcal{Z}^v_i = \text{ViT}(o_i)$. These visual features are later cross-attended~\cite{vaswani2017attention} with 32 learnable queries embedding $\mathcal{Q}_i \in \mathbf{R}^{32 \times 768}$ which is self-attended with text embedding $\mathcal{W}$ of the instruction first to obtain the instruction-aware image queries $\mathcal{Q'}_i$~\cite{dai2023instructblip}. These queries are fed to the LLM after a linear projection $W$ as the image tokens $\mathcal{H}^v_i = \mathcal{Q'}_i W$.

\subsubsection{\emph{Navigation System Prompt.}}
To inform LLM of the orientation of each candidate's view, we inject the direction information into the navigation prompt in structure input format "Candidate $i$, facing $a_i$, \{direction\}", shown in Figure~\ref{fig:prompt}. Moreover, we introduce special tokens \PredSty{\texttt{<IMG>}}, \PredSty{\texttt{</IMG>}}, \PredSty{\texttt{<INST>}} and \PredSty{\texttt{</INST>}} to insert images tokens and instructions into the prompt. Furthermore, we generate 10K navigational reasoning data from the R2R training set~\cite{anderson2018r2r} and perform instruction-tuning to the Q-former and the projection layer on the prediction tokens, using its original auto-regressive training objective, detail is discussed in section \ref{sec:training}. Up to now, a complete VLM has been built and can generate navigation reasoning with a standard LLM decoding process, shown in the bottom right of Figure~\ref{fig:architecture}.

\begin{figure}
    \begin{minipage}{\columnwidth}\vspace{0mm}
    \centering
    \begin{tcolorbox}
You are navigating in an indoor environment given the instruction: \PredSty{\texttt{<INST>}}\{instruction\}\PredSty{\texttt{</INST>}};

The navigable locations are listed below: \{

\quad"Candidate 1, facing $a_1$ degree, front" : \PredSty{\texttt{<IMG>}}\{image\_tokens\}\PredSty{\texttt{</IMG>}};

\quad"Candidate 2, facing $a_2$ degree, right" : \PredSty{\texttt{<IMG>}}\{image\_tokens\}\PredSty{\texttt{</IMG>}};

\quad...\}; 
  
  Please choose the next direction.
    \end{tcolorbox}
    \caption{Navigation system prompt for \ours.}
    \label{fig:prompt}
    \end{minipage}
\end{figure}

\subsubsection{\emph{VLM Latents as Visual-Linguistic Representation.}}\label{sec:global_policy}
For an LLM $f_\phi(\cdot)$ parameterized by $\phi$, we extract the LLM Latents $\mathcal{H'}^{v} = \{f_\phi(\mathcal{H}^v_i)\}^N_{i=1}, \mathcal{H'}^{l} = f_\phi(\mathcal{W})$ as Visual-Linguistic Representation for Navigation Policy. For encoder-decoder based LLMs, we retrieve the hidden representation of the image tokens and instruction tokens from the last Transformer encoder layer. For decoder-only LLMs, we obtain the latents from the last decoder layer. Specifically, the 32 image tokens for each view are merged into a single token through a multi-layer perception $\mathcal{\bar{H'}}^{v} = \text{MLP}(\mathcal{H'}^{v})$, shown in Figure~\ref{fig:architecture}.

\subsection{Graph Based Navigation Policy}
We identify the key difficulty of fine-tuning LLMs as VLN agents lies in the LLMs' inadequate comprehension of spatial structures, coupled with their limited ability to model the agent's long-range experiences during the navigation process. Therefore, we harness a topological graph-based navigation policy~\cite{chen2022duet} for effective action planning. The topological graph is maintained on the fly and served as a memorization mechanism to trace the navigation experience. \ours\ choose the next step from the entire constructed topological graph, enabling effective planning and back-tracing to unvisited nodes when a wrong path is taken. We introduce the graph-based policy in the following sections.

\subsubsection{\emph{Node Embedding.}} 
The graph memory consists of visited nodes and adjacent unexplored nodes along the trajectory. All the candidate views from each visited node are average-pooled to represent this node, while each underexplored node is represented by the partial pooling of corresponding views from all its adjacent visited nodes in the trajectory. Each view is represented by the summation of its visual features $\mathcal{\bar{H'}}^{v}$, its directional embedding $E^d$ representing the location of each node and step embedding $E^s$ representing the traverse order of the agent's current episode. The step embedding of the unexplored nodes is 0 and a `stop' node is added to the graph memory to denote a stop action.
A multi-layer transformer is implemented to model the spatial relation between nodes:
\begin{equation}
    \mathcal{V} = \text{SelfAttn}\left( \frac{1}{M}\sum^M_{i=1}\left(\mathcal{\bar{H'}}^{v}_i + E^d_i + E^s_i\right)\right),
\end{equation}
where $M$ is the number of views representing the node.

\subsubsection{\emph{Cross-Modal Encoding.}} 
The navigation graph constructs at step $t$ is denoted as $\mathcal{G}_t = \{\mathcal{V}_t, \mathcal{E}_t\}, \mathcal{G}_t \subset \mathcal{G}$. The node embeddings $\mathcal{V}_t$ are sent to a multi-layer cross-modal transformer to model the relationship between instructions and nodes. Specifically, the node embeddings are first cross-attended with the instructions encoded by the LLM, then go through a graph-aware self-attention (GASA), which considers both distances and visual similarities between nodes to enhance contextual understanding:
\begin{equation}
    \text{GASA}(\mathcal{V}) = \text{Softmax}\left( \frac{\mathcal{V}W_q(\mathcal{V}W_k)^T}{\sqrt{d}}+A(\mathcal{E}_t)\right)\mathcal{V}W_v,
\end{equation}
where $A(\mathcal{E}_t)$ represents the spatial affinity matrix, comprised of pairwise L2 distances among all observed nodes.

\subsubsection{\emph{Global Action Prediction.}} We employ a two-layer feed-forward network to process the output node representations of the GASA to generate an action score. The agent selects the node with the highest score as the target, and follows the shortest path in the graph memory to control to the selected node. Note that we mask the scores for visited nodes to encourage agent exploration following \cite{chen2022duet}.

\subsection{Multi-stage Learning for Action and Reasoning}\label{sec:training}
We perform a two-stage training to learn action prediction and navigation reasoning generation for LLM. In the first stage, we initialize the model from InstructBLIP~\cite{dai2023instructblip} after visual instruction-tuning on academic-task-oriented VQA datasets. We follow the same training schema to only finetune the Q-former with a frozen LLM and vision encoder on the collected navigation reasoning data, shown as the yellow blocks in Figure~\ref{fig:architecture}. In the second stage, we connect the pretrained VLM with the downstream navigation policy and only finetune the policy network with frozen VLM, shown as the red blocks in Figure~\ref{fig:architecture}.

\subsubsection{\emph{Data Acquisition and Curation.}}

To train VLM with navigational reasoning ability, we propose an automatic data generation pipeline with GPT-4V. We discard historical modeling for VLM and consider the situation when spanning the agent at the intermediate steps along the ground truth trajectory. We asked GPT-4V to determine the next step toward completing the instruction based on the current observation of the surroundings and relevant landmarks. We define the single-step navigation reasoning trace as describing the immediate environment and specifying the direction or action that will be taken to proceed. Details of the prompt are in the appendix.

\begin{figure}
    \centering
    \includegraphics[width=.99\linewidth]{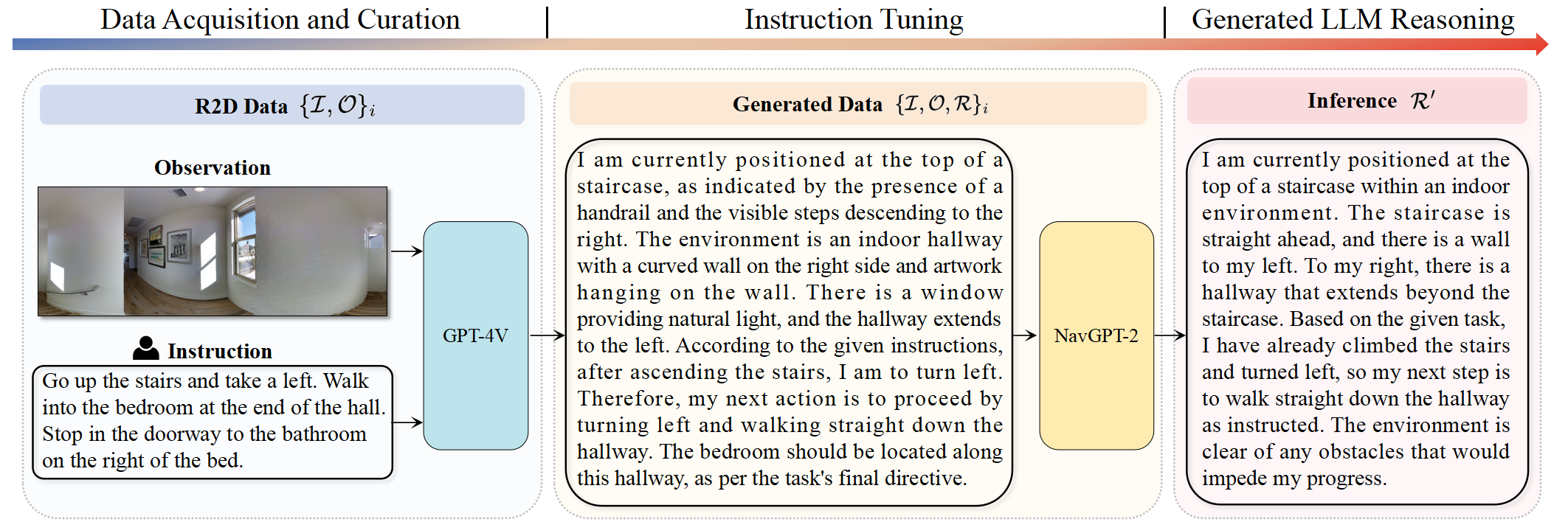}
    \caption{Data generation pipeline and visual instruction tuning on navigation reasoning data. $\{\mathcal{I}, \mathcal{O}\}$ denotes the instruction-observation pairs on the R2R trajectories. $\mathcal{R}$ is the generated reasoning from GPT-4V, $\mathcal{R}'$ is the generated reasoning from \ours.}
    \label{fig:data_generation}
\end{figure}

We randomly select 10k intermediate steps from the trajectory in the R2R~\cite{anderson2018r2r} training set, using the equirectangular projected panoramic image centring at the agent's heading direction as the image input for GPT-4V, shown in Figure~\ref{fig:data_generation}.

\subsubsection{\emph{Policy learning with DAgger.}}
When fine-tuning the downstream navigation policy network, we follow previous work to combine Behaviour cloning and DAgger loss~\cite{ross2011dagger}:
\begin{equation}
    \mathcal{L}_{\text{BC}} = -\sum^T_{t=1}\log{\pi(v^*_t|\mathcal{W}, \mathcal{G}_t)},\quad \mathcal{L}_{\text{DAG}} = -\sum^T_{t=1}\log{\pi(\tilde{v}^*_t|\mathcal{W}, \tilde{\mathcal{G}}_t)},
\end{equation}

where $v^*_t$ indicates the ground truth action, $\tilde{v}^*_t$ denotes the pseudo label determined by the shortest path toward the destination from the partial graph $\tilde{\mathcal{G}}_t$ generated by the agent through on-policy action sampling. The overall loss function is given by $\mathcal{L} = \lambda \mathcal{L}_{\text{BC}} + \mathcal{L}_{\text{DAG}}$, where $\lambda$ is a balancing factor.

\section{Experiments}


\subsubsection{\emph{Evaluation Metrics.}}
We adopt a comprehensive set of navigation metrics to evaluate performance~\cite{anderson2018r2r}, including Trajectory Length (TL), which measures the average path length in meters; Navigation Error (NE), the average distance between the final and target locations; Success Rate (SR), the percentage of paths with NE less than 3 meters; Oracle Success Rate (OSR), the success rate under an ideal stop policy; Success Rate penalized by Path Length (SPL)~\cite{anderson2018evaluation}, which combines success with efficiency considerations; Normalized Dynamic Time Warping (NDTW)~\cite{ilharco2019general}, assessing the fidelity between predicted and annotated paths; and Success weighted by Normalized Dynamic Time Warping (SDTW), a composite metric rewarding both navigation success and path fidelity.

\subsection{Implementation Details}
We build \ours{} based on InstructBLIP~\cite{dai2023instructblip} and exploit four variations of LLMs, including FlanT5-XL (3B), FlanT5-XXL (11B), Vicuna-7B and Vicuna-13B~\cite{chung2022scaling, vicuna2023}. All models apply the same vision encoder (ViT-g/14~\cite{fang2023eva}), and all parameters of the vision encoder and LLMs are kept frozen during the entire training process.
In stage one, we initialize the model from pretrained InstructBLIP checkpoints and train the Q-former for 200K steps with a batch size of 8. The AdamW optimizer~\cite{loshchilov2018decoupled} is employed and configured with $\beta_1=0.9, \beta_2=0.999$ and a weight decay of 0.05. To optimize learning efficiency, a linear warmup strategy is applied to the learning rate for the first 1,000 steps, gradually increasing it from $10^{-8}$ to $10^{-5}$, which is then followed by a cosine decay to a minimum learning rate of 0. In stage two, we freeze the pretrained VLM from stage one and finetune the downstream policy network with a batch size of 2 and a learning rate of $10^{-5}$. Our best model is trained on the combination of the R2R and the synthetic data from PREVALENT~\cite{hao2020prevalent}. All experiments are conducted on a single NVIDIA A100 GPU.

\subsection{Comparison with State of the Art}

We compare the single-run performance on the R2R dataset with previous SOTA methods in Table \ref{tab:r2r_full}. Specifically, we categorized them into distinct categories:

\begin{table*}[t]
\centering
\caption{Comparison of single-run performance on R2R dataset. \ours\ surpass all pervious methods incorporating LLMs and eliminate the gap between SOTA methods on the same training scale. $\dag$: Methods that use additional visual data than MP3D.
}
\resizebox{\textwidth}{!}{
\definecolor{Gray}{gray}{0.94} 

\begin{tabular}{lccccccccc>{\columncolor{Gray}}c>{\columncolor{Gray}}cccc>{\columncolor{Gray}}c>{\columncolor{Gray}}c} 

\toprule
\midrule
\multicolumn{1}{c}{\multirow{2}{*}{Methods}} & \multicolumn{1}{c}{\multirow{2}{*}{\parbox{1cm}{\vspace*{1ex}Freeze \\ \vspace*{1.5ex} LLM}}} & 
\multicolumn{5}{c}{Val Seen} & 
\multicolumn{5}{c}{Val Unseen} & 
\multicolumn{5}{c}{Test Unseen} \\ 

\cmidrule(r){3-7} 
\cmidrule(r){8-12} 
\cmidrule(r){13-17} 

\multicolumn{2}{c}{} & 
\multicolumn{1}{c}{TL} & 
\multicolumn{1}{c}{NE$\downarrow$} & 
\multicolumn{1}{c}{OSR$\uparrow$} & 
\multicolumn{1}{c}{SR$\uparrow$} & 
\multicolumn{1}{c}{SPL$\uparrow$} & 
\multicolumn{1}{c}{TL} & \multicolumn{1}{c}{NE$\downarrow$} & \multicolumn{1}{c}{OSR$\uparrow$} & \multicolumn{1}{c}{SR$\uparrow$} & \multicolumn{1}{c}{SPL$\uparrow$} & 
\multicolumn{1}{c}{TL} & \multicolumn{1}{c}{NE$\downarrow$} & \multicolumn{1}{c}{OSR$\uparrow$} & \multicolumn{1}{c}{SR$\uparrow$} & \multicolumn{1}{c}{SPL$\uparrow$} \\
\midrule
\midrule
Human    
& --
& -- & -- & -- & -- & -- 
& -- & -- & -- & -- & --
& 11.85 & 1.61 & 90 & 86 & 76 \\
\midrule
Seq2Seq~\cite{anderson2018r2r}
& --
& 11.33 & 6.01 & 53 & 39 & -- 
& 8.39 & 7.81 & 28 & 21 & -- 
& 8.13 & 7.85 & 27 & 20 & -- \\
RCM~\cite{wang2019reinforced}
& --
& 10.65 & 3.53 & 75 & 67 & -- 
& 11.46 & 6.09 & 50 & 43 & -- 
& 11.97 & 6.12 & 50 & 43 & 38 \\
Speaker Follower~\cite{fried2018speaker}
& --
& -- & 3.36 & 74 & 66 & -- 
& -- & 6.62 & 45 & 36 & -- 
& 14.82 & 6.62 & - & 35 & 28  \\
EnvDrop~\cite{tan2019envdrop}
& --
& 11.00 & 3.99 & -- & 62 & 59
& 10.70 & 5.22 & -- & 52 & 48
& 11.66 & 5.23 & 59 & 51 & 47 \\
\midrule
\rowcolor{Cerulean!20}\multicolumn{17}{l}{\emph{VLN Specialists with Vision-Language-Action Pretraining:}}\\
PREVALENT~\cite{hao2020prevalent}
& --
& 10.32 & 3.67 & -- & 69 & 65
& 10.19 & 4.71 & -- & 58 & 53
& 10.51 & 5.30 & 61 & 54 & 51 \\
AirBert~\cite{guhur2021airbert}$\dag$
& --
& 11.09 & 2.68 & -- & 75 & 70
& 11.78 & 4.10 & -- & 62 & 56 
& 12.41 & 4.13 & -- & 62 & 57 \\
\vlnbert~\cite{hong2020recurrent}
& --
& 11.13 & 2.90 & -- & 72 & 68
& 12.01 & 3.93 & -- & 63 & 57
& 12.35 & 4.09 & 70 & 63 & 57 \\
MARVAL~\cite{kamath2022marval}$\dag$
& --
& 10.60 & 2.99 & -- & 73 & 69
& 10.15 & 4.06 & -- & 65 & 61
& 10.22 & 4.18 & 67 & 62 & 58 \\
HAMT~\cite{chen2021hamt}
& --
& 11.15 & 2.51 & -- & 76 & 72
& 11.46 & 2.29 & -- & 66 & 61
& 12.27 & 3.93 & 72 & 65 & 60 \\
HOP+~\cite{qiao2023hop+}
& --
& 11.31 & 2.33 & -- & 78 & 73 
& 11.76 & 3.49 & -- & 67 & 61
& 12.67 & 3.71 & -- & 66 & 60 \\
BEVBert~\cite{an2022bevbert}
& --
& 13.56 & 2.17 & \textbf{88} & \textbf{81} & {74}
& 14.55 & 2.81 & 84 & 75 & 64
& 15.87 & 3.13 & 81 & 73 & 62 \\
DUET+ScaleVLN~\cite{wang2023scaling}$\dag$
& --
& 11.90 & {2.16} & {87} & {80} & \textbf{75}
& 12.40 & {2.34} & \textbf{87} & \textbf{79} & \textbf{70}
& 14.27 & {2.73} & \textbf{83} & \textbf{77} & \textbf{68} \\
\midrule\midrule
\rowcolor{Cerulean!20}\multicolumn{17}{l}{\emph{Baseline:}}\\
DUET~\cite{chen2022duet}
& --
& 12.32 & 2.28 & 86 & 79 & 73 
& 13.94 & 3.31 & 81 & 72 & 60
& 14.73 & 3.65 & 76 & 69 & 59 \\
\rowcolor{Orchid!20}\quad w/o local branch
& --
&-- & -- & -- & -- & --
&12.96 & 3.67 & -- & 68 & 59
&13.08 & 3.93 & -- & 67 & 58 \\
\midrule
\rowcolor{Cerulean!20}\multicolumn{17}{l}{\emph{Language Models zero-shot VLN:}}\\
NavGPT (GPT-4)~\cite{zhou2024navgpt}
& \cmark
& -- & -- & -- & -- & --
& 11.45 & 6.46 & 42 & 34 & 29
& -- & -- & -- & -- & -- \\
MapGPT (GPT-4)~\cite{chen2024mapgpt}
& \cmark
& -- & -- & -- & -- & --
& -- & 6.92 & 58 & 39 & 26
& -- & -- & -- & -- & -- \\
DiscussNav (GPT-4)~\cite{long2023discuss}
& \cmark
& -- & -- & -- & -- & --
& 9.69 & 5.32 & 61 & 43 & 40
& -- & -- & -- & -- & -- \\
\midrule
\rowcolor{Cerulean!20}\multicolumn{17}{l}{\emph{Langage Models with/as VLN Specialists:}}\\
NavCoT (LLaMA2-7B)~\cite{lin2024navcot}
& \xmark
& 10.08 & 6.46 & 48 & 41 & 38
& 9.95 & 6.26 & 48 & 40 & 37
& -- & -- & -- & -- & -- \\
LangNav (LLaMA2-7B)~\cite{pan2023langnav}$\dag$
& \xmark
& -- & -- & -- & 56 & --
& -- & -- & -- & 46 & --
& -- & -- & -- & -- & -- \\
NaviLLM (Vicuna-7B)~\cite{zheng2023towards}
& \xmark
& -- & -- & -- & -- & --
& 12.81 & 3.51 & -- & 67 & 59
& 13.21 & 3.71 & -- & 68 & \textbf{60} \\
\rowcolor{Orchid!20}$\text{\ours}_{\text{FlanT5-XL}}$ (ours, 1.5B) 
& \cmark
& 13.02 & 3.34 & 74 & 69 & 62
& 13.68 & 3.37 & 74 & 68 & 56
& -- & -- & -- & -- & -- \\
\quad w/ PREVALENT
& \cmark
& 12.44 & 2.97 & 80 & 73 & 65
& 12.81 & 3.33 & 79 & 70 & 59
& 13.51 & 3.40 & 77 & 71 & \textbf{60} \\
\rowcolor{Orchid!20}$\text{\ours}_{\text{FlanT5-XXL}}$  (ours, 5B)
& \cmark
& 13.08 & 2.98 & 79 & \textbf{74} & \textbf{65}
& 13.25 & 3.18 & 80 & 71 & 60
& -- & -- & -- & -- & -- \\
\quad w/ PREVALENT
& \cmark
%
& 14.13 & 2.84 & \textbf{83} & \textbf{74} & 63
& 14.01 & 2.98 & \textbf{84} & \textbf{74} & \textbf{61}
& 14.74 & 3.33 & \textbf{80} & \textbf{72} & \textbf{60} \\
\midrule
\bottomrule
\end{tabular}}
\label{tab:r2r_full}
\end{table*}

-- \emph{\textbf{VLN Specialists with Vision-Language-Action Pretraining}}~\cite{hao2020prevalent, guhur2021airbert, hong2020recurrent, kamath2022marval, chen2021history, qiao2023hop+, chen2022duet, an2022bevbert, wang2023scaling}: These methods are initialized from general vision-language models~\cite{tan2019lxmert, li2020oscar} and incorporate VLN-tailored pretraining with auxiliary tasks such as masked language modeling (MLM)~\cite{devlin2018bert}, masked region classification (MRC)~\cite{lu2019vilbert}, and single-step action prediction (SAP)~\cite{chen2021history} before fine-tuning on downstream VLN tasks.

-- \emph{\textbf{Zero-shot Methods}}~\cite{zhou2024navgpt, long2023discuss, chen2024mapgpt}: Methods that apply GPT-4 to zero-shot VLN using different textual inputs and prompting strategies.

-- \emph{\textbf{Methods finetuning LLMs}}~\cite{pan2023langnav, zheng2023towards, lin2024navcot}: Methods that finetune LLMs on VLN data with alternative modification on multimodality LLMs combining VLN specialized model designs.

-- \emph{\textbf{Baseline}}: We construct a baseline method based on the global action branch of DUET~\cite{chen2022duet}, referring as DUET (w/o local branch) in Table~\ref{tab:r2r_full}. This model shares the same architecture for the action planning policy network as \ours{} and trains on the same amount of data. The only difference lies in the language branch, where we harness the LLM's latent and the baseline adopt a 12-layer transformer initialized from LXMERT~\cite{tan2019lxmert}. Only a 2\% SR drop is observed on the test when cutting off the local branch for DUET, demonstrating the global branch as the dominating branch during navigation, facilitating \ours\ to simplify the navigation policy.

As shown in Table~\ref{tab:r2r_full}, we list our model with a similar size\footnote{Our model is smaller (1.5B and 5B) than original FlanT5 models (3B and 11B) as we only utilize the LLM encoder during navigation.} as previous LLM-based methods, $\text{\ours}_{\text{FlanT5-XXL}}$ (1.5B) and $\text{\ours}_{\text{FlanT5-XXL}}$ (5B). Our 5B model outperforms NaviLLM~\cite{zheng2023towards} (7B) by 4\% SR on test split while still maintaining the language capacity to generate self-explained navigation reasoning. \ours{} far exceeding NavGPT~\cite{zhou2024navgpt} and surpasses all the zero-shot methods, since those methods rely on extensive prompt engineering and struggle to model navigation history. Compared to the baseline methods, \ours\ bypass it by 5\% SR and 2\% SPL on the test split even if we do not incorporate with VLN pertaining.
The current SOTA method~\cite{wang2023scaling} is achieved by scaling up the training environment for DUET with HM3D~\cite{ramakrishnan2021hm3d} and Gibson~\cite{xia2018gibson}, besides the original 61 scenes in MP3D~\cite{chang2017matterport3d}. When considering the same training scale in MP3D, our method beats the original DUET by $3\%$ in SR on the test unseen split and is comparable with metric map-based method~\cite{an2022bevbert} which incorporates an extra depth sensor to construct Bird’s-Eye-View (BEV) perception. 

\subsection{Navigational Reasoning Generation}
In Figure~\ref{fig:reasoning} we show the navigational reasoning generated by \ours\ during navigation. \ours\ can construct a comprehensive perception of the surroundings, as shown in the example on the left, \ours\ recognizes the fireplace, dining room, hallway, and their relative locations. Moreover, it can reason about the navigation progress and identify associated sections of the instruction inferring the next step, and can even infer the expected observation. 

\begin{table*}[th]
\centering
\caption{Human study on \ours\ generated navigational reasoning.}
\resizebox{0.7\textwidth}{!}{
\begin{tabular}{lccc}
\midrule
\multicolumn{1}{c}{\textbf{Methods}}
& \textbf{Accuracy} & \textbf{ Informativeness} & \textbf{ Rationality} \\
\midrule
GPT-4V & 2.31 & 2.95 & 2.34 \\
$\text{\ours}_{\text{FlanT5-XL}}$ & 1.66 & 1.93 & 1.78 \\

\midrule
\end{tabular}
}

\label{tab:human_study}
\end{table*}

A human study is conducted in Table~\ref{tab:human_study} on 30 held-out samples to evaluate the accuracy and informativeness of VLM reasoning. Specifically, we engaged 10 volunteers to evaluate the navigational reasoning generated by \ours, focusing on its accuracy, informativeness, and rationality. The accuracy assessment, with a scoring range from 0 (Entirely incorrect) to 3 (Completely accurate), considered the precision in describing the surrounding environment, the correct recognition of navigation progress, and the appropriateness of the next action's planning. Informativeness was judged on a scale from 0 (Not informative) to 3 (Highly informative), based on the completeness of the environmental description provided. Rationality was also rated from 0 (Entirely irrational) to 3 (Completely rational), evaluating the correctness of the action planned by \ours.

As shown in Table~\ref{tab:human_study}, \ours\ scored 1.66 on Accuracy, 1.93 on Informativeness, and 1.78 on Rationality, indicating the quality of generated reasonings is acceptable given the full mark is 3. Noticeably the GPT-4V scored 2.31, 2.95, and 2.34 respectively, demonstrating an effective way to generate navigational reasoning data. 

\subsection{The Effect of Data Amount}
In Table~\ref{tab:data_efficiency} we initialize DUET from LXMERT and compare the model performance when finetuning 10\%, 50\%, and full R2R training data. 
 \ours\ outperforms all DUET variants in SR on the validation unseen split, and it reaches the same performance as DUET with full R2R data when feeding with 50\% less data, showcasing the data efficiency of utilizing LLMs latent as the vision-language representation and the benefits for downstream navigation policy learning.
\begin{table*}[bt]
\centering
\caption{Comparison of different scales of training data.}
\resizebox{0.9\textwidth}{!}{
\definecolor{Gray}{gray}{0.94}
\begin{tabular}{lcccc>{\columncolor{Gray}}c>{\columncolor{Gray}}cccc>{\columncolor{Gray}}c>{\columncolor{Gray}}c}
\toprule
\multicolumn{1}{c}{\multirow{2}{*}{Methods}} & 
\multicolumn{1}{c}{\multirow{2}{*}{\#}} &
\multicolumn{5}{c}{Val Seen} &
\multicolumn{5}{c}{Val Unseen} \\

\cmidrule(r){3-7}
\cmidrule(r){8-12}

\multicolumn{2}{c}{} &
\multicolumn{1}{c}{TL} &
\multicolumn{1}{c}{NE$\downarrow$} &
\multicolumn{1}{c}{OSR$\uparrow$} &
\multicolumn{1}{c}{SR$\uparrow$} &
\multicolumn{1}{c}{SPL$\uparrow$} &
\multicolumn{1}{c}{TL} & \multicolumn{1}{c}{NE$\downarrow$} & \multicolumn{1}{c}{OSR$\uparrow$} & \multicolumn{1}{c}{SR$\uparrow$} & \multicolumn{1}{c}{SPL$\uparrow$} \\
\midrule
\rowcolor{Cerulean!20}\multicolumn{12}{l}{\emph{10\% R2R training data:}} \\
DUET
& 1
& 12.75 & 6.18 & 53.97 & 44.47 & 38.43
& 13.08 & 5.78 & 58.19 & 48.19 & 39.93 \\
$\text{\ours}_{\text{FlanT5-XL}}$
& 2
& 14.46 & 6.13 & 57.59 & 45.64 & 36.80
& 13.66 & 5.21 & 62.37 & 52.02 & 41.75 \\
\midrule
\rowcolor{Cerulean!20}\multicolumn{12}{l}{\emph{50\% R2R training data:}} \\
DUET
& 3
& 13.46 & 4.65 & 68.07 & 56.51 & 49.49
& 13.57 & 4.41 & 70.50 & 59.90 & 50.11 \\
$\text{\ours}_{\text{FlanT5-XL}}$
& 4
& 12.95 & 3.99 & 68.85 & 61.51 & 52.55
& 14.01 & 3.98 & 71.90 & 63.30 & 51.83 \\
\midrule
\rowcolor{Cerulean!20}\multicolumn{12}{l}{\emph{100\% R2R training data:}} \\
DUET
& 5
& 12.38 & 3.62 & 73.36 & 66.31 & 60.13
& 13.20 & 4.07 & 71.95 & 63.90 & 54.83 \\
$\text{\ours}_{\text{FlanT5-XL}}$
& 6
& 13.02 & 3.34 & 74.24 & 69.44 & 61.72
& 13.68 & 3.37 & 74.37 & 67.52 & 56.01 \\
\midrule
\end{tabular}
}

\label{tab:data_efficiency}
\end{table*}

\subsection{Cross Dataset Generalization Ability}
We evaluate the generalization ability of \ours\ in two aspects: generalize to free-form language instructions and to various unseen environments.

\noindent\textbf{\emph{Generalize to Free-form Language Instructions.}} 
To assess \ours's comprehension of various forms of language instruction, we evaluate the zero-shot performance of \ours\ on the RxR dataset. The RxR dataset is characterized by its instructions of finer granularity, detailing rich landmarks and encompassing longer trajectories. 
As shown in Table~\ref{tab:rxr}, \ours\ outperforms DUET by $3.67\%$ in SR on RxR (English) unseen split, demonstrating our VLM-based method improves the out-of-domain language understanding capabilities.

\noindent\textbf{\emph{Generalize to Unseen Environments.}} 
We highlight that generalization challenges in unseen environments stem from the biases in connectivity graphs of training environments and visual differences in new scenes. Therefore, we assess \ours's zero-shot performance in HM3D by sampling 1000 trajectories from ScaleVLN~\cite{wang2023scaling} using Habitat Simulator rendered images, which offer environments visually deviant and structurally distinct graphs from MP3D. As shown in Table~\ref{tab:rxr}, \ours\ significantly outperforms DUET. We hypothesize this improvement is due to the projection of visual features into the same LLM hidden space as language, leading to a more robust alignment across environments.

\begin{table*}[t]
\centering
\caption{Comparison of zero-shot performance on RxR and HM3D.}
\resizebox{0.9\textwidth}{!}{
\definecolor{Gray}{gray}{0.94}
\begin{tabular}{llccc>{\columncolor{Gray}}c>{\columncolor{Gray}}cccc>{\columncolor{Gray}}c>{\columncolor{Gray}}c}
\toprule
\multicolumn{1}{c}{\multirow{2}{*}{Methods}} & 
\multicolumn{1}{c}{\multirow{2}{*}{\#}} &
\multicolumn{5}{c}{RxR-EN} &
\multicolumn{5}{c}{HM3D} \\

\cmidrule(r){3-7}
\cmidrule(r){8-12}

\multicolumn{2}{c}{} &
\multicolumn{1}{c}{nDTW$\uparrow$} &
\multicolumn{1}{c}{sDTW$\uparrow$} &
\multicolumn{1}{c}{OSR$\uparrow$} &
\multicolumn{1}{c}{SR$\uparrow$} &
\multicolumn{1}{c}{SPL$\uparrow$} &
\multicolumn{1}{c}{TL} &
\multicolumn{1}{c}{NE$\downarrow$} &
\multicolumn{1}{c}{OSR$\uparrow$} &
\multicolumn{1}{c}{SR$\uparrow$} &
\multicolumn{1}{c}{SPL$\uparrow$} \\

\midrule
DUET
& 1
& 37.77 & 17.39 & 44.54 & 25.07 & 19.65  
& 20.27 & 6.60 & 42.70 & 25.60 & 13.32 \\
$\text{\ours}_{\text{FlanT5-XXL}}$
& 2
& 38.50 & 19.24 & 48.56 & \textbf{28.75} & \textbf{22.36} 
& 17.96 & 4.91 & 69.80 & \textbf{47.20} & \textbf{27.99} \\
\midrule
\end{tabular}
}

\label{tab:rxr}
\end{table*}

\subsection{Ablation Study}
We ablate the core design choices applied in this paper, including the effect of incorporating a navigation-specific policy model, pretraining the Q-former with reasonings and leveraging different LLMs in \ours{}.

\noindent\textbf{\emph{Effect of Policy Network.}}
In Table~\ref{tab:ablation_variants}, we study the necessity of applying a navigation-specific policy model in \ours{}. To achieve this, we remove all the visual-language cross-attention layers in the Q-former and policy network and use only a single graph-aware self-attention layer followed by a single feed-forward layer to predict the action (Model\#2). By doing so, we force the LLM to take over the visual-textual based decision-making to exploit its navigational capability. It is clear from the drastic drop in performance that a frozen LLM is incapable of inferring effective representations that indicate a correct action. Although tuning the LLM should improve the results, we can see from previous work that tuning a LLaMA2-7B model on the full R2R data and GPT-4-augmented data only leads to a 43\% SR in Val Unseen~\cite{pan2023langnav}, still far behind \ours{}, implying that this approach might be infeasible with existing LLMs.

\noindent\textbf{\emph{Effect of Pretraining with Reasonings.}}
Following the above, we also highlighted the navigational reasoning abilities that \ours{} unleashed from frozen LLMs in Figure~\ref{fig:data_generation}. Additionally, we can see from Model\#3 of Table~\ref{tab:ablation_variants} that the pretraining of Q-former on reasonings brings slight improvement to the success rates of the model. It is expected that such information, containing rich spatial descriptions and visual landmarks, facilitates the Q-former to produce better textual representations of the multi-view observations.
\begin{table*}[htp]
\centering
\caption{Effect of navigation policy network and pretrained Q-former for reasoning.}
\resizebox{\textwidth}{!}{
\definecolor{Gray}{gray}{0.94}
\begin{tabular}{llccc>{\columncolor{Gray}}c>{\columncolor{Gray}}cccc>{\columncolor{Gray}}c>{\columncolor{Gray}}c}
\toprule
\multicolumn{1}{c}{\multirow{2}{*}{Methods}} & 
\multicolumn{1}{c}{\multirow{2}{*}{\#}} &
\multicolumn{5}{c}{Val Seen} &
\multicolumn{5}{c}{Val Unseen} \\

\cmidrule(r){3-7}
\cmidrule(r){8-12}

\multicolumn{2}{c}{} &
\multicolumn{1}{c}{TL} &
\multicolumn{1}{c}{NE$\downarrow$} &
\multicolumn{1}{c}{OSR$\uparrow$} &
\multicolumn{1}{c}{SR$\uparrow$} &
\multicolumn{1}{c}{SPL$\uparrow$} &
\multicolumn{1}{c}{TL} & \multicolumn{1}{c}{NE$\downarrow$} & \multicolumn{1}{c}{OSR$\uparrow$} & \multicolumn{1}{c}{SR$\uparrow$} & \multicolumn{1}{c}{SPL$\uparrow$} \\
\midrule
$\text{\ours}_{\text{FlanT5-XL}}$
& 1
& 13.02 & 3.34 & 74.24 & 69.44 & 61.72
& 13.68 & 3.37 & 74.37 & 67.52 & 56.01 \\
w/o policy model
& 2
& 25.57 & 8.03 & 68.85 & 25.27 & 13.92
& 26.70 & 8.03 & 69.60 & 21.46 & 10.23 \\
w/o pre-trained Q-former for reasoning
& 3
& 12.29 & 3.67 & 71.79 & 67.58 & 60.62
& 13.04 & 3.75 & 73.82 & 66.75 & 57.10 \\
\midrule
\end{tabular}
}

\label{tab:ablation_variants}
\end{table*}

\noindent\textbf{\emph{Effect of Different LLMs.}}
Technically, \ours{} can incorporate a wide range of different LLMs, however, their performance on VLN might not scale with the model size.
We compare four variations of LLMs in \ours{}, shown in Table~\ref{tab:ablation_llm}. For the encoder-decoder model FlanT5, a $3.79\%$ increment in SR is observed on the Val Unseen split when the model size is increased from XL (Model\#1, 1.5B) to XXL (Model\#2, 5B). But for the decoder-only models Vicuna (Model\#3 and Model\#4), although they are larger in size than FlanT5-XXL, their performance is much worse. These results in fact align with the findings in InstructBLIP~\cite{dai2023instructblip} which shows that the FlanT5-based model is superior at multi-choice questions while Vicuna-based models are generally better at open-ended generation tasks due to the difference in their training data. As a result, it is unsurprising that FlanT5-based \ours{} performs better in the multi-choice setting of VLN. Furthermore, we hypothesize that the full attention in FlanT5's encoder is more effective in addressing the alignment task between visual and instruction sequences in VLN than the causal attention in decoder-only Vicuna. We will leave a detailed investigation of this problem for future work.
\begin{table*}[t]
\centering
\caption{Comparison of different LLMs.}
\resizebox{0.85\textwidth}{!}{
\definecolor{Gray}{gray}{0.94}
\begin{tabular}{llccc>{\columncolor{Gray}}c>{\columncolor{Gray}}cccc>{\columncolor{Gray}}c>{\columncolor{Gray}}c}
\toprule
\multicolumn{1}{c}{\multirow{2}{*}{Methods}} & 
\multicolumn{1}{c}{\multirow{2}{*}{\#}} &
\multicolumn{5}{c}{Val Seen} &
\multicolumn{5}{c}{Val Unseen} \\

\cmidrule(r){3-7}
\cmidrule(r){8-12}

\multicolumn{2}{c}{} &
\multicolumn{1}{c}{TL} &
\multicolumn{1}{c}{NE$\downarrow$} &
\multicolumn{1}{c}{OSR$\uparrow$} &
\multicolumn{1}{c}{SR$\uparrow$} &
\multicolumn{1}{c}{SPL$\uparrow$} &
\multicolumn{1}{c}{TL} & \multicolumn{1}{c}{NE$\downarrow$} & \multicolumn{1}{c}{OSR$\uparrow$} & \multicolumn{1}{c}{SR$\uparrow$} & \multicolumn{1}{c}{SPL$\uparrow$} \\

\midrule
$\text{\ours}_{\text{FlanT5-XL}}$
& 1
& 13.02 & 3.34 & 74.24 & 69.44 & 61.72
& 13.68 & 3.37 & 74.37 & 67.52 & 56.01 \\
$\text{\ours}_{\text{FlanT5-XXL}}$
& 2
& 13.08 & 2.98 & 79.43 & 73.65 & 65.25
& 13.25 & 3.18 & 79.61 & 71.31 & 60.07 \\
$\text{\ours}_{\text{Vicuna-7B}}$
& 3
& 11.85 & 4.85 & 63.37 & 53.57 & 46.83
& 12.29 & 4.86 & 65.56 & 53.77 & 45.26 \\
$\text{\ours}_{\text{Vicuna-13B}}$
& 4
& 11.89 & 5.09 & 61.31 & 52.01 & 45.95 
& 13.16 & 5.63 & 60.11 & 48.28 & 40.14 \\
\midrule
\end{tabular}
}

\label{tab:ablation_llm}
\end{table*}

\section{Conclusion}

In this work, we strive to eliminate the gap between LLMs-based agents and VLN-specialised agents, while maintaining the interpretative intrinsic of LLMs to generate navigational reasonings during navigation. Through comprehensive experimentation, we highlight the critical aspects of integrating LLMs with downstream navigation policy networks. It is demonstrated that the Vision-Language Model (VLM) latent serves as a superior and more efficient visual-linguistic representation, enabling policy networks to learn better alignment between vision, language, and action. Our approach offers a scalable framework to leverage the broad language comprehension capabilities of LLMs, paving the way for the development of versatile navigation agents capable of interacting with humans and understanding free-form human intentions with greater efficacy.

\section*{Acknowledgements}
We thank all the reviewers for their valuable comments and suggestions. Yicong Hong wants to thank NVIDIA for the Academic Hardware Grant that provided GPU support for this project. This project is supported by the University of Adelaide’s Centre for Augmented Reasoning (CAR).

%
%
\bibliographystyle{splncs04}
\bibliography{main}

\appendix
\newpage
\section*{Supplementary Material for \ours}

Section~\ref{sec:duet} provides additional details for DUET as it is the main comparison with \ours. The prompt for GPT-4V used in the data generation pipeline and additional experiment results are described in Section~\ref{sec:gpt4v} and Section~\ref{sec:addition}. Section~\ref{sec:limit} illustrates the limitation of \ours\ with the discussion of future directions. Finally, Section~\ref{sec:broader} discusses the broader impacts of our work.

\section{DUET Revisit}\label{sec:duet}

\ours\ exploit the similar design adapted from Dual-scale Graph Transformer (DUET)~\cite{chen2022duet} as the downstream navigation policy. It includes a text encoder to encode instructions, a global and a local branch to enable coarse-scale and fine-scale cross-modal reasoning. 

\subsection{Text Embedding and Visual Embedding}
For the text encoder, DUET utilizes a 12-layer transformer initialized from LXMERT~\cite{tan2019lxmert}. For visual embedding, the visual observation at each node is 36 view images from 12 horizontal directions times 3 vertical directions. To distinguish these nodes, a directional embedding $E^{ang}$ of the absolute angle for each view is added to the visual feature $\mathcal{Z}^{v}$ extracted by the vision encoder. Moreover, since DUET inputs all 36 view images to construct the spatial observation for the model, the navigable adjacent nodes are only observed at a few view images, denoted as navigable views. A navigable embedding $E^{nav}$ is added to the visual features. The final visual embedding is sent to a 2 layers transformer to encode the spatial relations between views and obtain the panoramic view embeddings:

\begin{equation}
    \mathcal{H}^{pano} = \text{SelfAttn}\left( \mathcal{Z}^{v} + E^{ang} + E^{nav}\right).
\end{equation}

On the contrary, \ours\ only inputs the navigable views, thus the directional embedding $E^{ang}$ and the navigable embedding $E^{nav}$ are removed in the downstream policy, instead we directly add the step embedding and location embedding before sending to the 2 layers transformer.

\subsection{DUET Local Branch}
\ours\ adopt the same navigation policy network architecture as the DUET global branch, discussed in \S 3.2\protect\footnote{Refer to section 3.2 in the main paper.}, so we omit the explanation of the global branch in DUET. In this section, we introduce the local branch of DUET. This branch performs action prediction based on the current node's instruction and egocentric observation. No graph information is provided besides the local observation.
\subsubsection{Local Visual Embedding}
Two types of location embedding are added to the panoramic view embedding $\mathcal{H}^{pano}$. The first type is the relative location of the current node to the starting node, to encode the long distance direction between nodes. The second type is each adjacent view to the current node, to encode egocentric directions such as "turn right".

\subsubsection{Local Cross-model Encoding}
The local branch utilizes a standard cross-modal transformer of 4 layers to model vision and language relations. During action prediction, a mask is set to the unnavigable views, and the action logits are only calculated for the navigable views at the current node.

\subsection{Dynamic Fusion}
The final action prediction of DUET is performed by dynamically fusing the action predicted by local and global branches. The local branch predicts actions within the adjacent nodes $\mathcal{V}^a_t$. It is incongruent with the action space used by the global branch, which chooses the next action from all nodes $\mathcal{V}_t$ in the constructed graph at step $t$. To reconcile this discrepancy, the local action scores $s^l_i$ are transformed encompassing options such as ``stop'' and $\mathcal{V}_t$, into a representation suitable for the global action space by summing up scores of visited nodes in $\mathcal{V}^a_t$ as a backtrack score $s_b$:

\begin{equation}
s^{l'}_{i} = 
\begin{cases} 
s_{\text{b}}, & \text{if } \mathcal{V}_{i} \in \mathcal{V}_{t} - \mathcal{V}^a_t, \\
s^{l'}_{i}, & \text{otherwise}.
\end{cases}
\end{equation}

This adjustment facilitates navigation toward other unexplored nodes not directly linked to the current node, necessitating the agent to retrace its steps through neighboring nodes that have previously been visited. The final navigation score is given by:

\begin{equation}
    s_i = \sigma_t s_i^g + (1 - \sigma_t) s^{l'}_{i},
\end{equation}
where $s_i^g$ is the logits from global branch, $\sigma_t$ is a learnable scalar for fusion.

\section{GPT-4V Prompt}\label{sec:gpt4v}

\begin{figure}
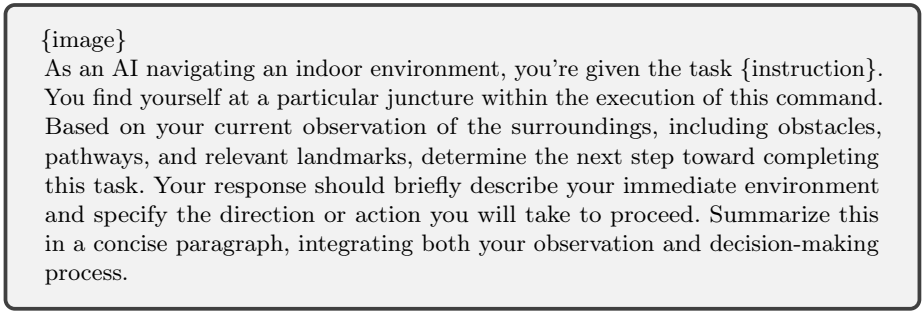

    \begin{minipage}{\columnwidth}\vspace{0mm}
    \centering
    \begin{tcolorbox}
\{image\}

As an AI navigating an indoor environment, you're given the task \{instruction\}.

You find yourself at a particular juncture within the execution of this command. Based on your current observation of the surroundings, including obstacles, pathways, and relevant landmarks, determine the next step toward completing this task. Your response should briefly describe your immediate environment and specify the direction or action you will take to proceed. Summarize this in a concise paragraph, integrating both your observation and decision-making process.

    \end{tcolorbox}
    \caption{Navigation reasoning generation prompt for GPT-4V.}
    \label{fig:gpt}
    \end{minipage}
\end{figure}

The prompt used for GPT-4V to generate navigation reasoning, discussed in section \S 3.3 is shown in Figure~\ref{fig:gpt}.

\section{Additional Rerults}\label{sec:addition}

In this section, we conduct additional experiments to illustrate the choice of navigation policy network for \ours\ and the effectiveness of LLM features. To align the same training schema of the navigation policy, we conduct the experiments for DUET initiating it from LXMERT without  VLN specialized pretraining.

\begin{table*}[t]
\centering
\caption{Comparison of single-run performance on R2R dataset.}
\resizebox{\textwidth}{!}{
\definecolor{Gray}{gray}{0.94} 

\begin{tabular}{lcccc>{\columncolor{Gray}}c>{\columncolor{Gray}}cccc>{\columncolor{Gray}}c>{\columncolor{Gray}}c} 

\toprule
\midrule
\multicolumn{1}{c}{\multirow{2}{*}{Methods}} & \multicolumn{1}{c}{\multirow{2}{*}{\#}} & 
\multicolumn{5}{c}{Val Seen} & 
\multicolumn{5}{c}{Val Unseen} \\

\cmidrule(r){3-7} 
\cmidrule(r){8-12} 

\multicolumn{2}{c}{} & 
\multicolumn{1}{c}{TL} & 
\multicolumn{1}{c}{NE$\downarrow$} & 
\multicolumn{1}{c}{OSR$\uparrow$} & 
\multicolumn{1}{c}{SR$\uparrow$} & 
\multicolumn{1}{c}{SPL$\uparrow$} & 
\multicolumn{1}{c}{TL} & \multicolumn{1}{c}{NE$\downarrow$} & \multicolumn{1}{c}{OSR$\uparrow$} & \multicolumn{1}{c}{SR$\uparrow$} & \multicolumn{1}{c}{SPL$\uparrow$} \\

\midrule
\midrule
{\emph{w/o Visual-Language-Action Pretrain:}}\\
DUET
& 1
& 12.38 & 3.62 & 73 & 66 & 60
& 13.20 & 4.07 & 72 & 64 & 55\\
\quad w/o local branch
& 2
& 11.43 & 3.50 & 74 & 67 & 62
& 12.08 & 4.08 & 71 & 62 & 54 \\
\quad w/ EVA-CLIP-g
& 3
& 12.64 & 3.73 & 73 & 66 & 60
& 14.27 & 4.07 & 72 & 63 & 54\\
\midrule
$\text{\ours}_{\text{FlanT5-XL}}$ (ours, 1.5B) 
& 4
& 13.02 & 3.34 & 74 & 69 & 62
& 13.68 & 3.37 & 74 & 68 & 56 \\
$\text{\ours}_{\text{FlanT5-XXL}}$  (ours, 5B)
& 5
& 13.08 & 2.98 & 79 & 74 & 65
& 13.25 & 3.18 & 80 & 71 & 60 \\
\bottomrule
\end{tabular}}
\label{tab:duet}
\end{table*}

\begin{figure}
    \centering
    \includegraphics[width=.85\linewidth]{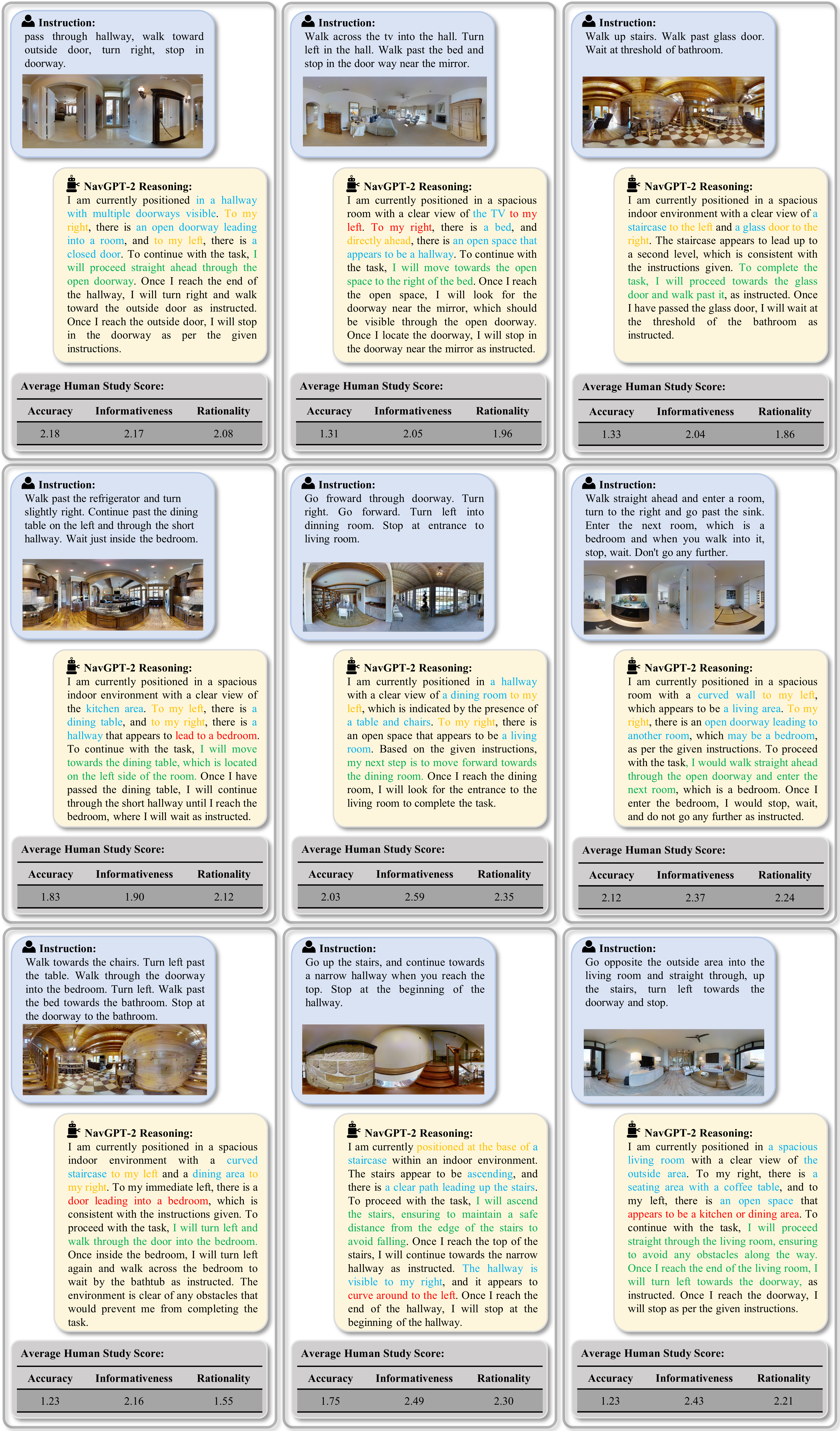}
    \caption{Qualitive Results for \ours. It can \textcolor{myblue}{correctly recognize object and scenes} and their \textcolor{myorange}{corresponding locations}, \textcolor{mygreen}{grouding the observation to the given instruction and plan the next step}. However, \textcolor{red}{hallucination of the non-existent object or misjudged the direction} is also observed.}
    \label{fig:qualitative}
\end{figure}

\subsection{Effect of Vision Encoder}

Because \ours\ exploits a stronger vision encoder~\cite{fang2023eva}, we conduct an ablation study on the original DUET to investigate the performance gain brought by the vision encoder. As shown in Table~\ref{tab:duet}, after switching the visual representation to the stronger vision feature same as \ours, little performance gain is observed for the DUET global branch (Model \# 3 compared to Model \# 2). We hypothesize this is due to the global branch for DUET performing vision-language alignment on a coarse scale, while the fine-grained alignment is performed in the local branch. Therefore, the main performance gain in \ours\ is not contributed by the stronger vision encoder but the better representation from LLM hidden.

\subsection{Effect of VLN Pretrain}
We consider the same training scale and the same training schema of DUET as \ours, without pertaining auxiliary VLN tasks and directly finetuning on the VLN dataset. Under the same training schema and scale of data, \ours\ performs significantly better than the original DUET, shown in Table~\ref{tab:duet}. This showcases the superiors of LLM features that enable the learning of cross-modality alignment in the downstream task when the visual feature is projected to the LLM's language space by the Q-former. Without VLN tailored pertaining, the performance of DUET significantly drops. We leave adding the pertaining process for the downstream navigation policy in future work.

\subsection{Additional Qualitive Results}

In this section, we present extra qualitative results in addition to \S 4.3. In Figure~\ref{fig:qualitative}, we present the navigational reasoning produced by \ours\ during navigation. \ours\ is capable of \textcolor{myblue}{forming a detailed understanding of its surroundings with objects and scenes} and \textcolor{myorange}{their corresponding orientations}. Furthermore, it adeptly \textcolor{mygreen}{reasons about the progress of navigation and correlates it with specific portions of the instruction}. Impressively, it is also able to predict expected observations, such as "appears to lead to a bedroom," based on the current visual inputs. This demonstrates \ours's ability not only to navigate but also to anticipate and interpret complex environments intelligently.

\section{Limitations and Future Work}\label{sec:limit}

Although \ours\ could generate navigation reasoning to some extent, it is hard to evaluate the effectiveness of these reasonings, since it is set as a single-step reasoning based on local observation and does not model the navigation history in the VLM. Instead, such history information is encoded in the downstream navigation policy. As a result, the consistency between navigation reasonings is underexplored. Moreover, the reasoning and action predicted by downstream navigation policy are not strictly synchronized in \ours, such synchronization could be done either explicitly by tuning LLM with the same supervision signal of action or by collaborating with the reasoning generation loss during fine-tuning the downstream policy network, we leave the synchronization to future work. Finally, the communicative capability of \ours\ is not evaluated in this work, we suggest investigating the communicative ability of LM-based VLN agents and the synchronization between their reasoning and actions as a future direction.

\section{Broader Effect}\label{sec:broader}

Our research endeavors to leverage Large Vision-Language Models (VLM) to develop VLN agents, while preserving the linguistic prowess of VLMs for explaining action predictions in natural language. We posit that the inherent communicative capability, commonsense knowledge, and broad linguistic comprehension of VLM constitute the cornerstone for creating instruction-following navigation agents with generalizability. \ours\ illuminates the reasonings of VLM throughout the navigation process explicitly and interpretably. Due to safety and ethical considerations, we currently conduct all experiments using the open-source Vision-and-Language Navigation dataset within a simulated environment, which ensures controlled agent behavior. Concurrently, we acknowledge that the potential practical application of this technology warrants further exploration, particularly in terms of action and reasoning synchronization, which remains an underexplored area. Notably, we observe the propensity of VLMs to hallucinate non-existent scenes or objects and fail to identify object directions, shown in Figure~\ref{fig:qualitative}, which is also a common issue within VLM research. Future investigations are essential to address how to harmonize VLM action and reasoning and to enhance the agent's ability to self-explain in a manner intelligible to humans, a critical consideration for ensuring safety in real-world applications.

%
%

\end{document}